\documentclass[pdflatex,sn-mathphys-num]{sn-jnl}

\usepackage{graphicx}%
\usepackage{multirow}%
\usepackage{amsmath,amssymb,amsfonts}%
\usepackage{amsthm}%
\usepackage{mathrsfs}%
\usepackage[title]{appendix}%
\usepackage{xcolor}%
\usepackage{textcomp}%
\usepackage{newunicodechar}
\newunicodechar{−}{\textminus}
\usepackage{manyfoot}%
\usepackage{booktabs}%
\usepackage{algorithm}%
\usepackage{algorithmicx}%
\usepackage{algpseudocode}%
\usepackage{listings}%
\usepackage[most]{tcolorbox} 
\usepackage[most]{tcolorbox}
\usepackage{multicol}   
\newcounter{mybox}
\usepackage{tabularx}
\usepackage{booktabs}
\usepackage[most]{tcolorbox}
\usepackage[table]{xcolor}
\usepackage{tabularx}
\usepackage{colortbl}
\usepackage{xcolor}
\usepackage{caption}
\usepackage{graphicx}
\usepackage{subcaption}
\usepackage{capt-of}
\usepackage{makecell}
\usepackage[section]{placeins} 
\definecolor{headerbg}{RGB}{250, 225, 205}
\usepackage[switch]{lineno} 
\newcolumntype{L}{>{\raggedright\arraybackslash}X}
\newtcolorbox{mhbox}[1][]{
  breakable,
  enhanced,
  colback=gray!12,      
  colframe=gray!12,     
  boxrule=0pt,
  left=1em, right=1em, top=0.6em, bottom=0.9em,
  before skip=10pt, after skip=10pt,
  overlay={
    \draw[line width=0.6pt, black]
      ([xshift=1em,yshift=-2.2em]frame.north west) -- ([xshift=-1em,yshift=-2.2em]frame.north east);
  },
  #1
}

\theoremstyle{thmstyleone}%
\theoremstyle{thmstyletwo}%

\theoremstyle{thmstylethree}%
\newcounter{snote}

\newcounter{edtab}

\usepackage{caption}
\DeclareCaptionLabelSeparator{pipe}{\ \textbar\ }
\makeatletter
\newcommand{\fakelabel}[2]{%
  \protected@write\@auxout{}{\string\newlabel{#1}{{#2}{\thepage}}}%
}
\makeatother

\begin{document}
\fakelabel{edtab:overview_dataset}{1}
\fakelabel{edtab:task_orders}{2}
\fakelabel{edtab:scale_params}{3}
\fakelabel{supply:icons}{1}
\fakelabel{supply:final_ap}{2}
\fakelabel{supply:paired_tests}{3}
\fakelabel{supply:forgetting_curve_ord5_order8}{4}
\fakelabel{supply:transition_shock_ord5_8}{5}
\fakelabel{supply:family_mapping}{6}
\fakelabel{supply:task-family_raw}{7}
\fakelabel{supply:training_time_backbone}{8}

\title[Article Title]{MedCL-Bench: Benchmarking stability–efficiency trade-offs and scaling in biomedical continual learning}


\author{\fnm{Min} \sur{Zeng}}

\author{\fnm{Shuang} \sur{Zhou}}

\author{\fnm{Zaifu} \sur{Zhan}}

\author*{\fnm{Rui} \sur{Zhang}}\email{ruizhang@umn.edu}

\affil[1]{\orgdiv{Division of Computational Health Sciences, Department of Surgery}, \orgname{University of Minnesota}, \orgaddress{\city{Minneapolis}, \postcode{55455}, \state{MN}, \country{USA}}}

\abstract{

Medical language models must be updated as evidence and terminology evolve, yet sequential updating can trigger catastrophic forgetting. Although biomedical NLP has many static benchmarks, no unified, task-diverse benchmark exists for evaluating continual learning under standardized protocols, robustness to task order and compute-aware reporting. We introduce \textbf{MedCL-Bench}, which streams ten biomedical NLP datasets spanning five task families and evaluates eleven continual learning strategies across eight task orders, reporting retention, transfer, and GPU-hour cost. 
Across backbones and task orders, direct sequential fine-tuning on incoming tasks induces catastrophic forgetting, causing update-induced performance regressions on prior tasks. Continual learning methods occupy distinct retention--compute frontiers: parameter-isolation provides the best retention per GPU-hour, replay offers strong protection at higher cost, and regularization yields limited benefit. Forgetting is task-dependent, with multi-label topic classification most vulnerable and constrained-output tasks more robust. MedCL-Bench provides a reproducible framework for auditing model updates before deployment.
}

\maketitle

\clearpage







Large language models (LLMs) are increasingly used to support biomedical question answering, evidence retrieval, relation extraction, and document-level classification. Yet biomedical knowledge is not static: new findings, revised clinical evidence, and evolving therapeutic guidance continually change what models should know. 
However, updating large models by full retraining is often computationally impractical, while repeatedly fine-tuning on new datasets can erode previously acquired capabilities through \emph{catastrophic forgetting}~\citep{mccloskey1989catastrophic,he2025harnessing}. The result is a tension between \emph{plasticity} (incorporating new knowledge) and \emph{stability} (preserving prior competencies) that is especially consequential in biomedical settings, where outdated or inconsistent behavior can directly undermine downstream research and decision-support.  
Continual learning (CL) has therefore emerged as a promising paradigm for enabling models to acquire new knowledge over time while mitigating degradation of prior competencies~\citep{DBLP:journals/tmlr/MendezE23}.

In the clinical domain, these challenges are sharpened by three practical constraints~\citep{kiyasseh2021clinical,lee2020clinical}. 
First, privacy and data-governance constraints often limit the pooling of raw patient records across institutions, creating persistent ``data walls'' between sites~\citep{guinney2018alternative, nmi_trust_ai_medicine_2024}. 
This motivates \emph{model-to-data} workflows in which models are updated sequentially across hospitals (e.g., Hospital A $\rightarrow$ Hospital B) by transferring parameters rather than sensitive primary data~\citep{bergquist2020model2data,mcmahan2017fedavg}. 
Second, clinical data are inherently sequential and longitudinal\citep{abulhusn2019ehr,huguet2020using}: patient information accrues over time through repeated encounters, tests, treatments, and evolving diagnoses, while labeled datasets and task definitions often emerge incrementally rather than as a single static corpus.
Third, clinical practice exhibits continual \emph{knowledge drift}: diagnostic standards, treatment guidelines, and pathogen profiles can change over time~\citep{sahiner2023datadrift,finlayson2021datasetshift,lasko2024transport}, requiring models to incorporate new evidence without repeated full retraining. 
These constraints translate directly into deployment risk. Sequential updates can introduce \emph{silent regressions}~\citep{finlayson2021datasetshift,kore2024driftdetection}: a model may improve on newly introduced data while degrading on previously validated capabilities. In safety-critical workflows, such regressions increase the verification burden for clinicians and can undermine the reliability of downstream decision-support, including tasks such as adverse-event monitoring or drug--drug interaction detection. This creates a practical need for rigorous evaluation of continual learning behavior under repeated updates.

Despite growing interest in both biomedical Natural Language Processing (NLP) and continual learning, several questions remain unresolved in realistic biomedical deployment settings. In particular, it remains unclear how severe catastrophic forgetting is in biomedical NLP, whether CL methods reliably mitigate forgetting and deliver meaningful benefits in biomedical NLP, which classes of CL strategies are most effective for biomedical tasks, whether findings from general-domain CL transfer to settings with specialized biomedical language and heterogeneous data distributions, whether larger backbones consistently improve performance or instead exhibit non-monotonic, backbone-dependent trade-offs, and how these methods compare in terms of training cost, parameter efficiency, and update-time overhead. Biomedical NLP still lacks a unified, task-diverse benchmark for CL under a standardized training and evaluation protocol to answer these questions rigorously.

\begin{figure}[ht]
    \centering
    \includegraphics[width=\linewidth]{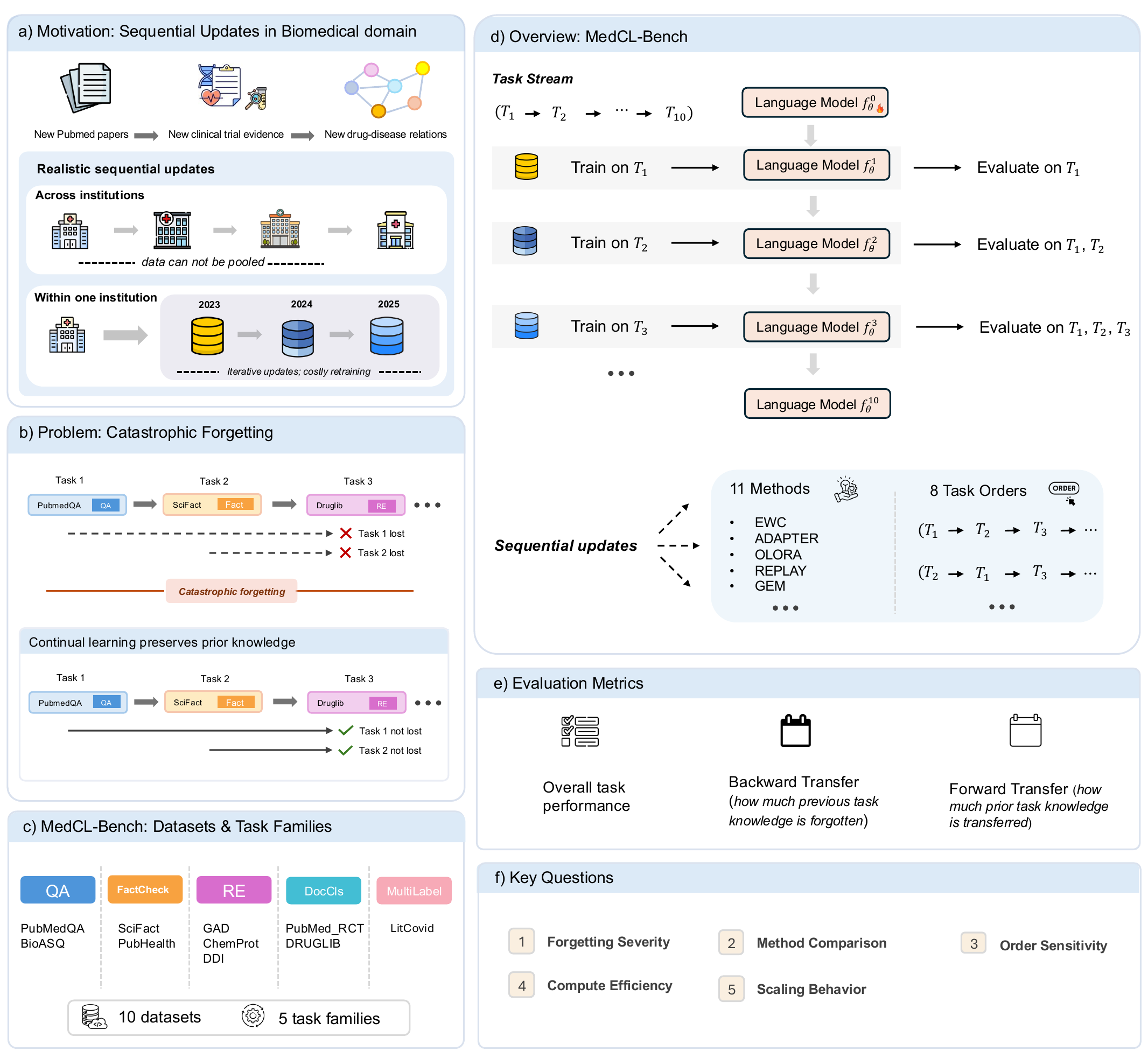}
    \caption{\textbf{Overview of MedCL-Bench.}
     \textbf{(a)} Biomedical knowledge and datasets evolve continuously (e.g., new literature and drug--disease relations), creating realistic sequential update streams---both across institutions (where data cannot be pooled) and within an institution over time.
    \textbf{(b)} Sequential sequential fine-tuning can overwrite previously acquired capabilities (catastrophic forgetting), whereas CL aims to retain prior knowledge while learning new tasks.
    \textbf{(c)} MedCL-Bench comprises ten biomedical NLP datasets grouped into five task families (QA, fact checking, relation extraction, document classification, and multi-label topic classification).
    \textbf{(d)} Benchmark workflow: a pretrained backbone is updated sequentially on a task stream under multiple task orders, and evaluated on all previously seen tasks after each stage.
    \textbf{(e)} CL metrics reported in this work: overall task performance (AP), backward transfer (BWT), and forward transfer (FWT).
    \textbf{(f)} Key questions addressed: forgetting severity, method comparison, order sensitivity, compute efficiency, and scaling/backbone dependence.
    Icons are sourced from Flaticon.com (full attributions in Supplementary Note~\ref{supply:icons}). 
    }
    \label{fig:medcl_overview}
\end{figure}

To address these gaps, we introduce \textbf{MedCL-Bench} (Fig.~\ref{fig:medcl_overview}), a unified continual learning benchmark for biomedical NLP. MedCL-Bench is designed to answer three practical questions: (i) how severe catastrophic forgetting is in biomedical NLP, (ii) which continual learning strategies offer the best trade-offs between end-of-stream performance and training cost, and (iii) how sensitive conclusions are to task order and backbone choice under realistic resource constraints. 
Specifically, MedCL-Bench provides a standardized benchmark suite and evaluation pipeline for controlled comparison of continual learning strategies in biomedical NLP.
First, we curate and standardize ten public biomedical NLP datasets into a standardized continual learning benchmark suite spanning five task families: biomedical question answering ( \textsc{PubMedQA},\textsc{BioASQ}), scientific fact checking (\textsc{SciFact}, \textsc{PubHealth}), relation extraction (\textsc{GAD}, \textsc{ChemProt}, \textsc{DDI}), document-level classification (\textsc{Pubmed\_RCT}, \textsc{DRUGLIB}), and multi-label topic classification of biomedical literature (\textsc{LitCovid}). 
Second, we evaluate sequential updates across eight pre-specified task orders with a unified preprocessing and evaluation protocol, enabling matched comparisons across continual learning strategies and backbone architectures. 
Third, we benchmark representative continual learning strategies---including naive fine-tuning, multi-task learning~\citep{caruana1997multitask}, regularization~\citep{kirkpatrick2017overcoming, zenke2017continual, aljundi2018memory}, rehearsal/gradient projection~\citep{lopez2017gradient, chaudhry2018efficient, mi2020continual}, generative replay~\citep{zhao2022prompt,zeng2024dirichlet,sun2019lamol}, and parameter-efficient adaptation~\citep{madotto2021continual,zeng2025task}---and report end-of-stream performance together with compute- and parameter-efficiency (including GPU-hour cost) to characterize stability--efficiency trade-offs. Finally, we release the full benchmark codebase—preprocessing, training, and evaluation—to enable reproducible comparisons and further research.


Using this benchmark, we further find that retention is not uniform across biomedical tasks. Tasks with more constrained output structure, such as multiple-choice question answering and multi-class relation extraction, are comparatively robust to sequential updates, whereas multi-label topic classification with overlapping label sets is markedly more vulnerable to forgetting. These patterns suggest that forgetting in biomedical continual learning depends not only on task difficulty, but also on task formulation and output structure.

\section*{Results}\label{sec2}
We evaluate continual learning on MedCL-Bench, a stream of ten public biomedical NLP datasets
spanning five task types: biomedical question answering, scientific fact checking,
relation extraction, document-level classification, and multi-label topic classification.  
A detailed overview of all datasets is provided in Extended Data Table~\ref{edtab:overview_dataset}.


Tasks are presented sequentially under eight randomized task orders (Extended Data Table~\ref{edtab:task_orders}).
Models are incrementally updated without access to future tasks. After each training stage, models are evaluated on all previously encountered tasks, enabling systematic assessment of both knowledge acquisition and retention.
Unless stated otherwise, results use the \textsc{T5-base} backbone; additional backbones are considered only in the scaling experiments.

Across all experiments, we report three continual learning metrics: average performance (AP; mean of per-task accuracies over all tasks after completing a task order), backward transfer (BWT; less negative indicates less forgetting), and forward transfer (FWT; less negative indicates better transfer)
.
Table~\ref{tab:main_results} summarizes AP/BWT/FWT for each method under each of the eight task orders.

\subsection*{Overall performance across methods}
Using Table~\ref{tab:main_results}, we highlight the main trends across methods.
Multi-task learning (MULTI) provides an empirical upper bound, achieving consistently high AP ($\sim$76\%) across orders when all tasks are jointly optimized.
Among continual learning methods, ADAPTER and TCL attain the strongest and most consistent AP (72.01--73.27\% and 69.75--71.37\%, respectively), closely approaching the upper bound.
GEM achieves higher AP but is more order-sensitive (66.83--73.69\% across orders), whereas REPLAY improves over naïve sequential fine-tuning (VANILLA) yet remains lower overall (58.05--63.55\%).

\textsc{VANILLA} denotes standard sequential fine-tuning with no CL mechanism. For forgetting, BWT values reveal a sharp separation between method families.
VANILLA shows severe forgetting with consistently negative BWT across all orders, whereas regularization baselines (EWC, L2) only partially mitigate forgetting and still yield negative BWT.
In contrast, memory-based approaches---especially GEM and REPLAY---substantially mitigate forgetting, with BWT values markedly closer to zero.

FWT is consistently negative across methods, indicating limited forward transfer. VANILLA forgets most (BWT down to $-57.69$), while replay/constraint methods improve retention (e.g., GEM: $-6.88$ to $-1.33$).
Notably, OLORA remains highly forgetting-prone (BWT $-43.92$ to $-29.33$), underscoring that stability requires explicit retention mechanisms, not only parameter-efficient updates.

\clearpage
\begin{center}
\captionof{table}{\textbf{Overall performance on MedCL-Bench.}
For each method, we report average task performance (AP), backward transfer (BWT),
and forward transfer (FWT) across ten biomedical NLP tasks and eight randomized task orders.
Higher is better; AP is highlighted in \textcolor{red}{red}.
}
\label{tab:main_results}
\includegraphics[width=\textwidth]{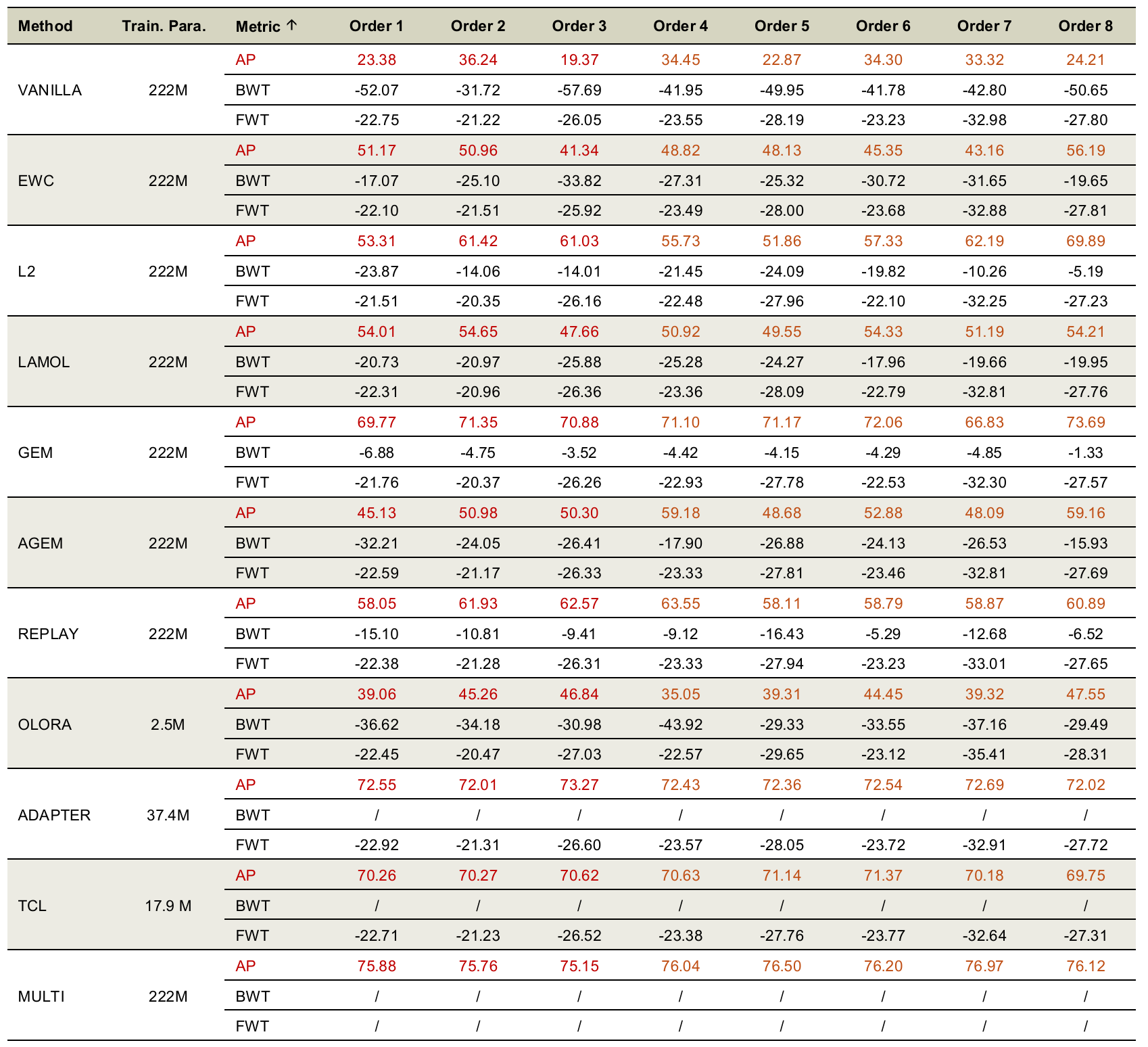}
\end{center}


\subsection*{Order robustness and statistical reliability}
\begin{figure}[ht]
    \centering
    \begin{subfigure}[t]{0.49\linewidth}
        \centering
        \includegraphics[width=\linewidth]{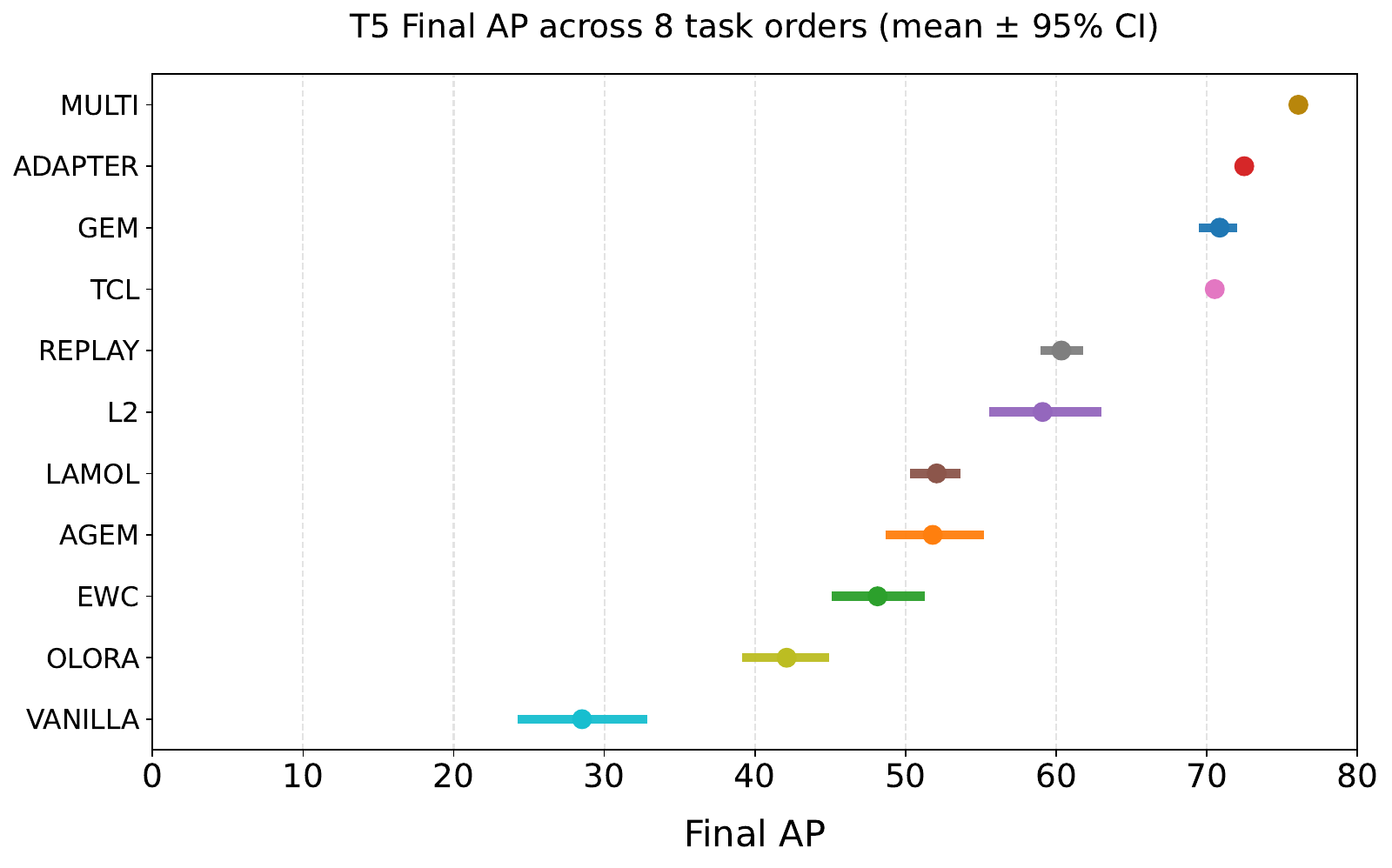}
        \caption{\textbf{Mean with 95\% bootstrap CI.} Final AP aggregated over 8 task orders.}
        \label{fig:order_robust_a}
    \end{subfigure}\hfill
    \begin{subfigure}[t]{0.49\linewidth}
        \centering
        \includegraphics[width=\linewidth]{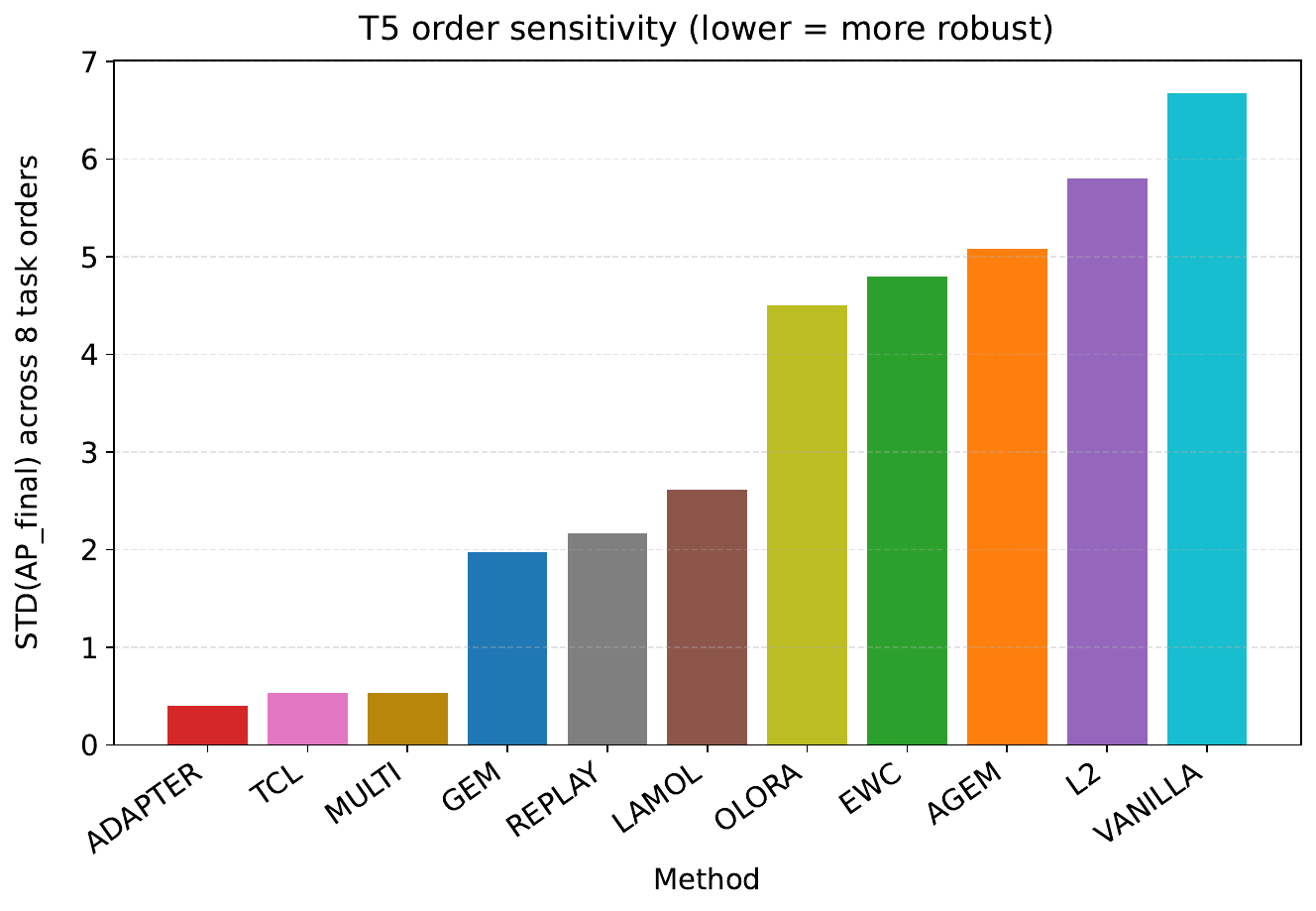}
        \caption{\textbf{Order sensitivity.} Standard deviation of final AP across orders (lower = more robust).}
        \label{fig:order_robust_b}
    \end{subfigure}

    \caption{\textbf{Order robustness and statistical reliability on MedCL-Bench (\textsc{T5-base}).}
    (a) Mean final AP across eight randomized task orders with 95\% bootstrap confidence intervals for the mean obtained by resampling task orders (n=8).
    (b) Order sensitivity measured as the standard deviation (s.d.) of final AP across orders (lower indicates stronger robustness).
    Together, these panels summarize both average performance and sensitivity to task-order permutations.}
    \label{fig:order_robustness}
\end{figure}

Figure~\ref{fig:order_robustness} quantifies sensitivity of final AP to task-order permutations.
Fig.~\ref{fig:order_robust_a} reports mean final AP across eight orders with 95\% bootstrap confidence intervals obtained by resampling task orders ($n{=}8$); MULTI provides an empirical upper bound with a narrow interval.


Among CL methods, ADAPTER and TCL show the highest mean AP with the tightest intervals, whereas GEM achieves comparable mean AP but with wider intervals, indicating residual order sensitivity. REPLAY improves over VANILLA but remains lower on average.

Complementing the CI-based view, Fig.~\ref{fig:order_robust_b} summarizes order sensitivity using the standard deviation of final AP across the eight orders.
ADAPTER, TCL, and MULTI show the smallest observed variability, whereas GEM and REPLAY exhibit intermediate sensitivity.
Several baselines (notably VANILLA and L2) have the largest standard deviations, indicating that their outcomes can vary substantially under different task permutations.
Per-order final AP values are shown in Supplementary Note~\ref{supply:final_ap}.
Treating task order as a matched block ($n{=}8$), paired exact sign-flip tests with Holm correction confirm that ADAPTER and TCL outperform REPLAY, whereas L2 does not (Supplementary Note~\ref{supply:paired_tests}).

\subsection*{Forgetting Dynamics Across Task Orders}\label{sec:forget_curves}
To understand how task order affects stability during training, we examine stage-wise trajectories.
Fig.~\ref{fig:order_dynamics_a} shows $AP_t$, the mean accuracy over tasks seen up to stage $t$, under four task orders (Orders~1--4); trajectories for Orders~5--8 are provided in Supplementary Note~\ref{supply:forgetting_curve_ord5_order8}.
This metric provides an intuitive view of the stability--plasticity trade-off: sharp drops indicate substantial forgetting and interference with previously learned tasks, whereas flat trajectories indicate robust retention.

Across all orders, VANILLA exhibits recurrent collapses as training progresses. Although the drops occur at different transition points depending on the permutation, the overall pattern remains consistent: performance on previously learned tasks deteriorates under long-horizon sequential updates and only partially recovers.

Memory-based methods provide the most consistent stabilization in these trajectories. In particular, REPLAY (experience replay) and GEM maintain comparatively smooth trajectories across orders, with markedly reduced drops when new tasks arrive. 
Notably, REPLAY interleaves a memory buffer of past samples during training, whereas GEM additionally enforces gradient constraints---projecting updates to avoid increasing loss on stored examples---at the cost of extra computation. 
Regularization-based approaches (EWC, L2) provide only partial protection and still exhibit noticeable degradations at several transitions.


\subsubsection*{Transition-level diagnosis of order sensitivity}
While the forgetting curves in Fig.~\ref{fig:order_dynamics_a} summarize the stage-wise evolution of $AP_t$, they do not identify which task switches cause abrupt changes.
We therefore analyze transition shock, defined as $\Delta AP_t = AP_{t+1}-AP_t$ (in percentage points), where $AP_t$ averages performance over all tasks observed up to stage $t$.
Fig.~\ref{fig:order_dynamics_b} reports transition shocks for Order~1--4; Order~5--8 are provided in Supplementary Note~\ref{supply:transition_shock_ord5_8}.


Across task orders, memory-based methods (REPLAY, GEM) and parameter-isolation methods (ADAPTER, TCL) tend to reduce the magnitude of negative transition shocks, whereas sequential fine-tuning (VANILLA) and regularization baselines (EWC, L2) exhibit larger drops at multiple switches, indicating stronger interference when new tasks are introduced.
This transition-level view aligns with the aggregate trends in Table~\ref{tab:main_results}: methods with larger and more frequent negative shocks typically show more negative BWT (greater forgetting).
Finally, while FWT captures transfer to a task \emph{before} it is trained, $\Delta AP_t$ reflects the net change \emph{after} learning the next task (acquisition minus interference), so strong FWT does not necessarily imply small transition shocks.
Taken together, these trajectory- and transition-level diagnostics show that method stability can depend on task permutations, motivating evaluation across multiple orders rather than relying on a single sequence.
We next examine whether forgetting differs systematically across task families.

\clearpage
\begin{figure}[t]
\centering
\begin{subfigure}[t]{\linewidth}
  \centering
  \includegraphics[width=\linewidth]{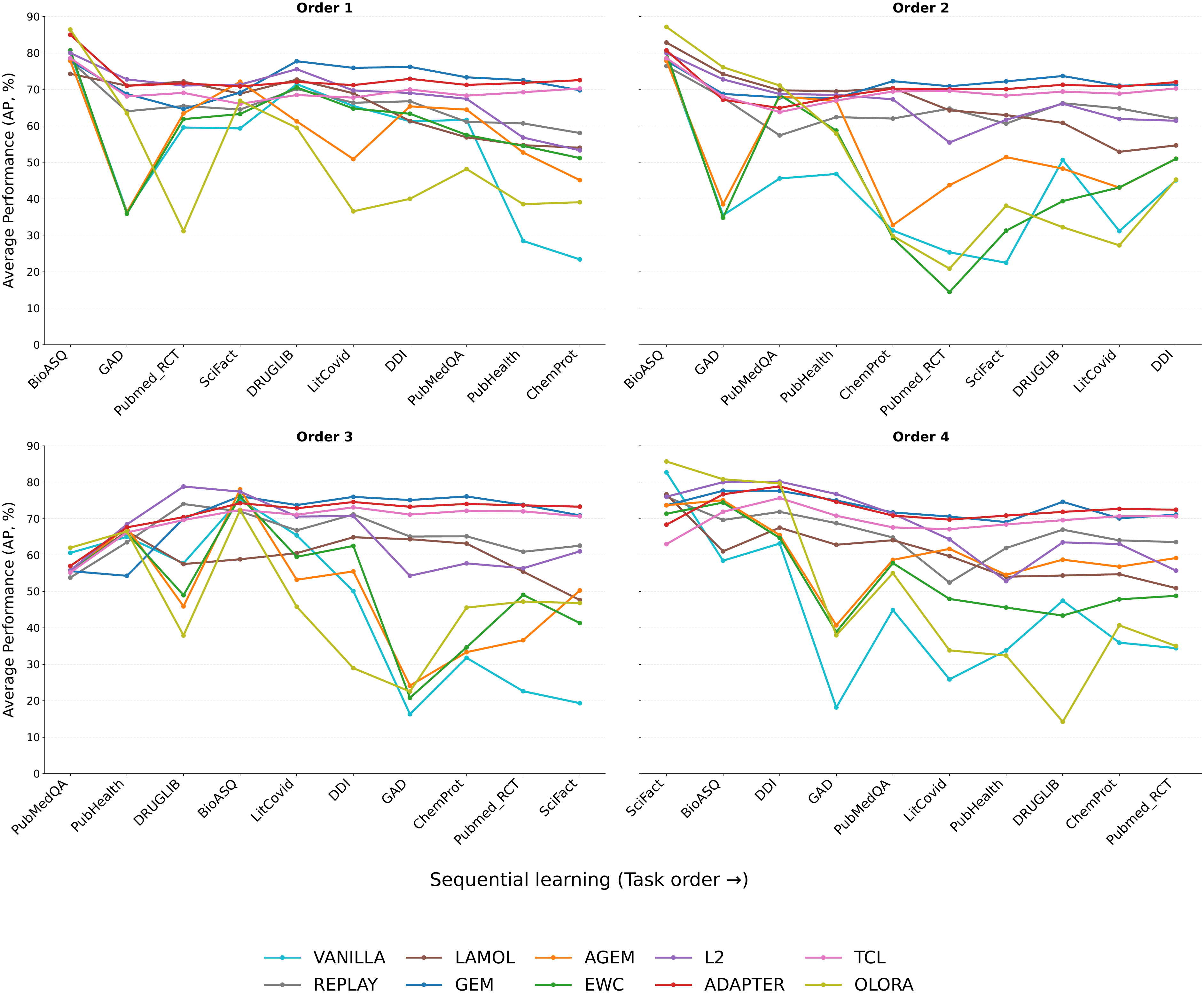}
  \caption{\textbf{Forgetting curves under four task orders.}
  Trajectories of seen-task average accuracy across stages for Orders~1--4.}
  \label{fig:order_dynamics_a}
\end{subfigure}
\caption{\textbf{Order-dependent forgetting dynamics in MedCL-Bench.}
\textbf{(a)} Forgetting curves. \textbf{(b)} Transition shock heatmaps (next page).}
\label{fig:order_dynamics}
\end{figure}

\begin{figure}[t]\ContinuedFloat
\centering
\begin{subfigure}[t]{\linewidth}
  \centering
  \begin{subfigure}[t]{0.49\linewidth}
    \centering
    \includegraphics[width=\linewidth]{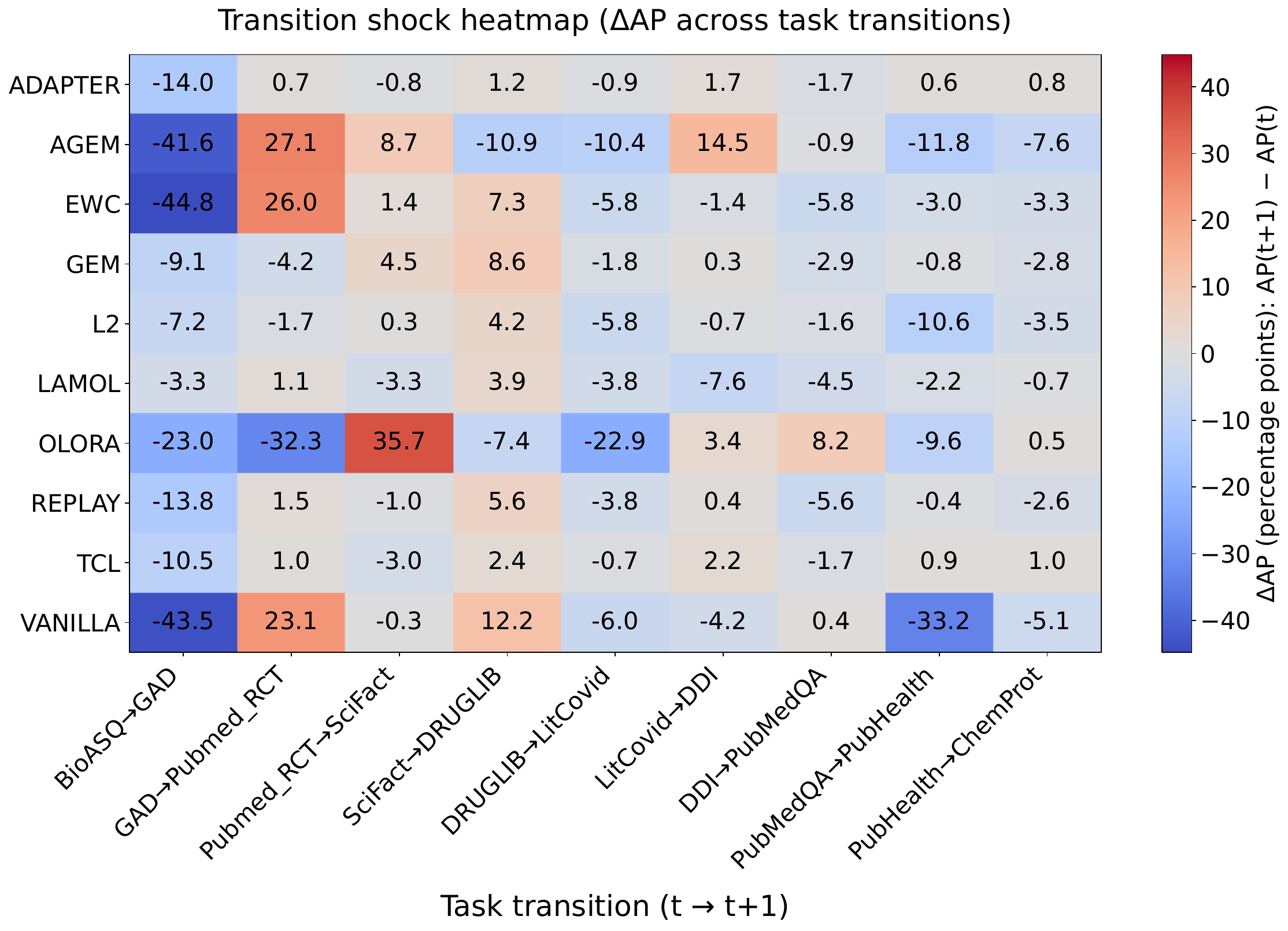}
    \caption*{Order~1}
  \end{subfigure}\hfill
  \begin{subfigure}[t]{0.49\linewidth}
    \centering
    \includegraphics[width=\linewidth]{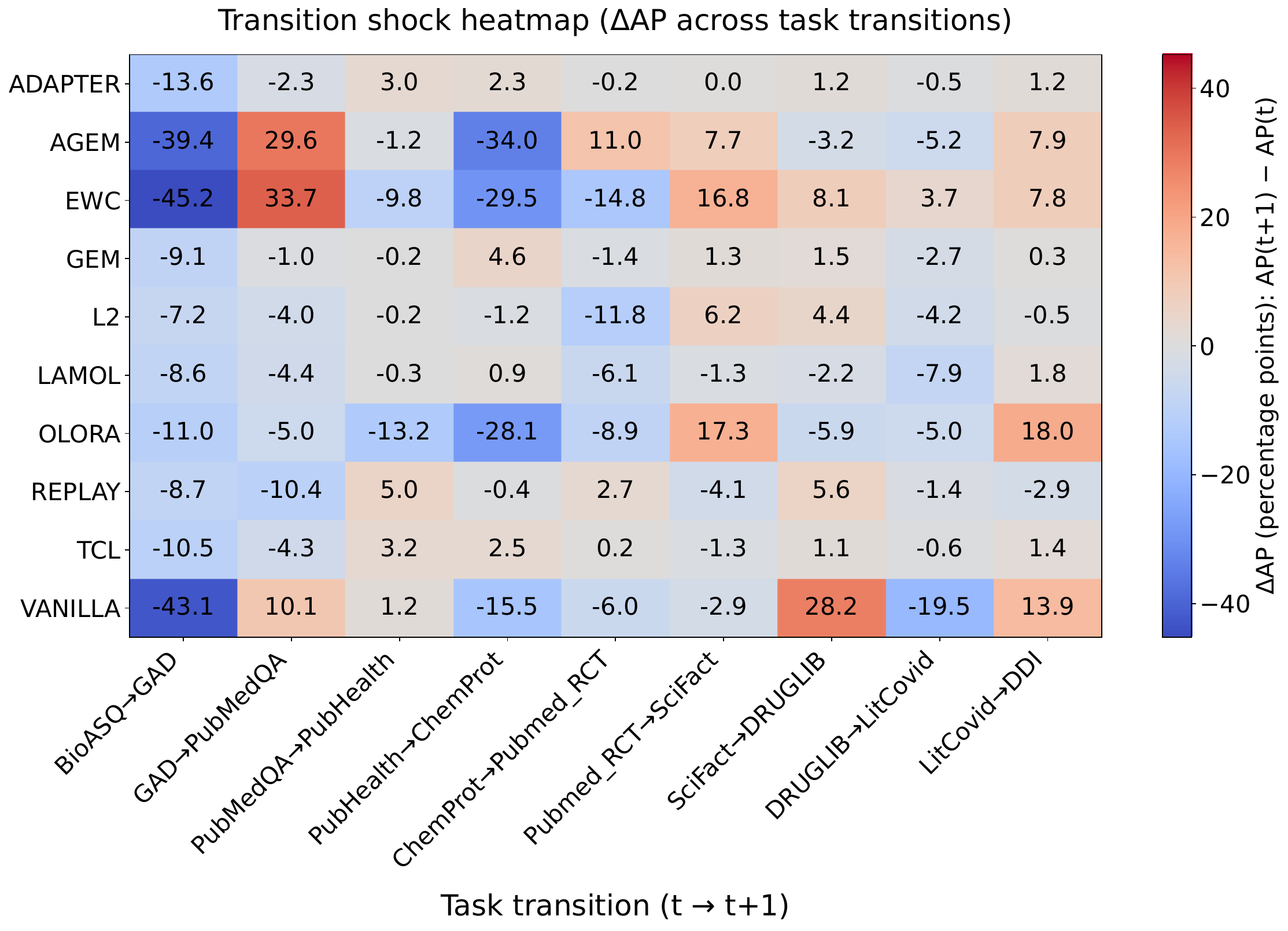}
    \caption*{Order~2}
  \end{subfigure}

  \vspace{0.6em}

  \begin{subfigure}[t]{0.49\linewidth}
    \centering
    \includegraphics[width=\linewidth]{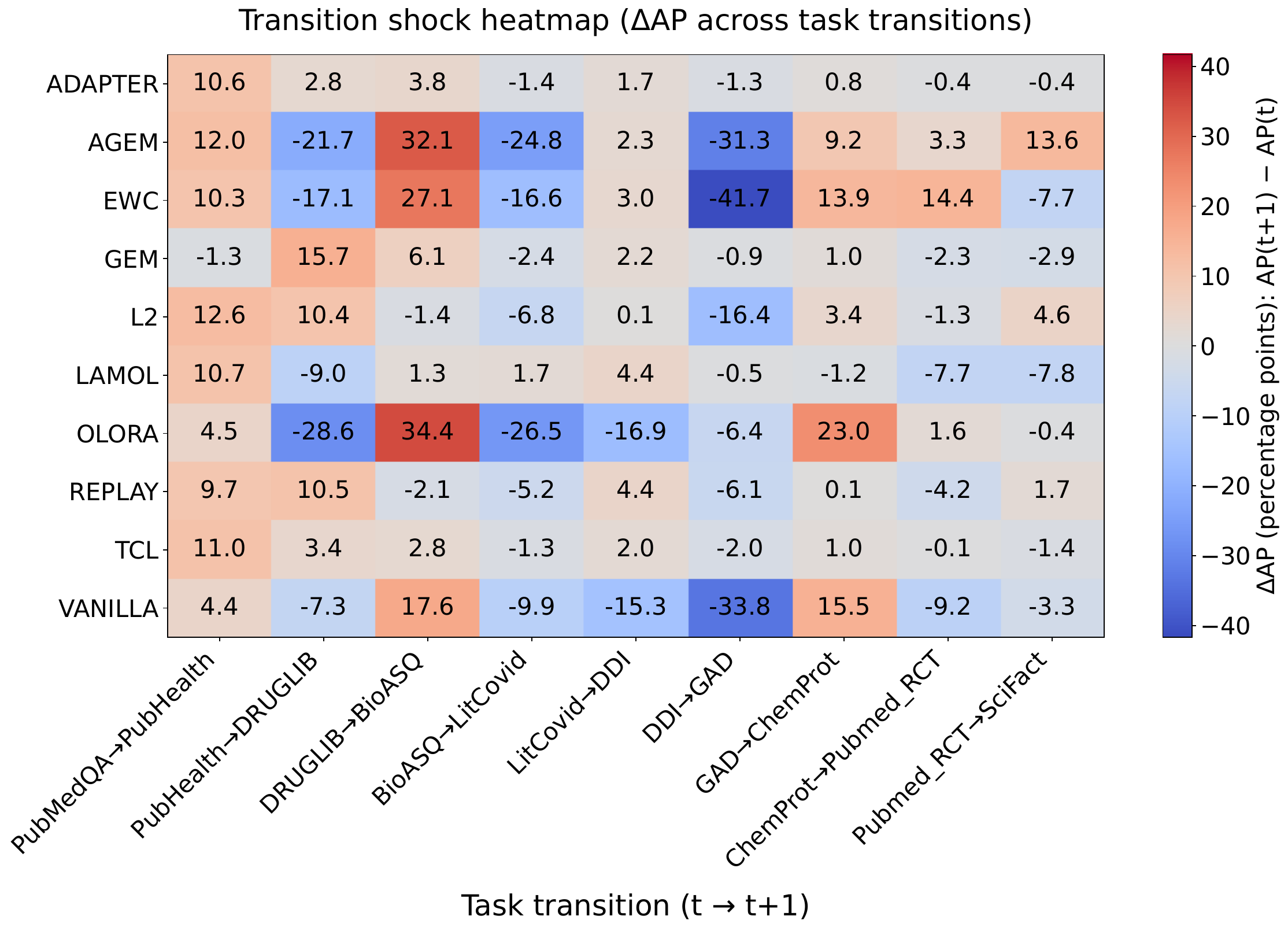}
    \caption*{Order~3}
  \end{subfigure}\hfill
  \begin{subfigure}[t]{0.49\linewidth}
    \centering
    \includegraphics[width=\linewidth]{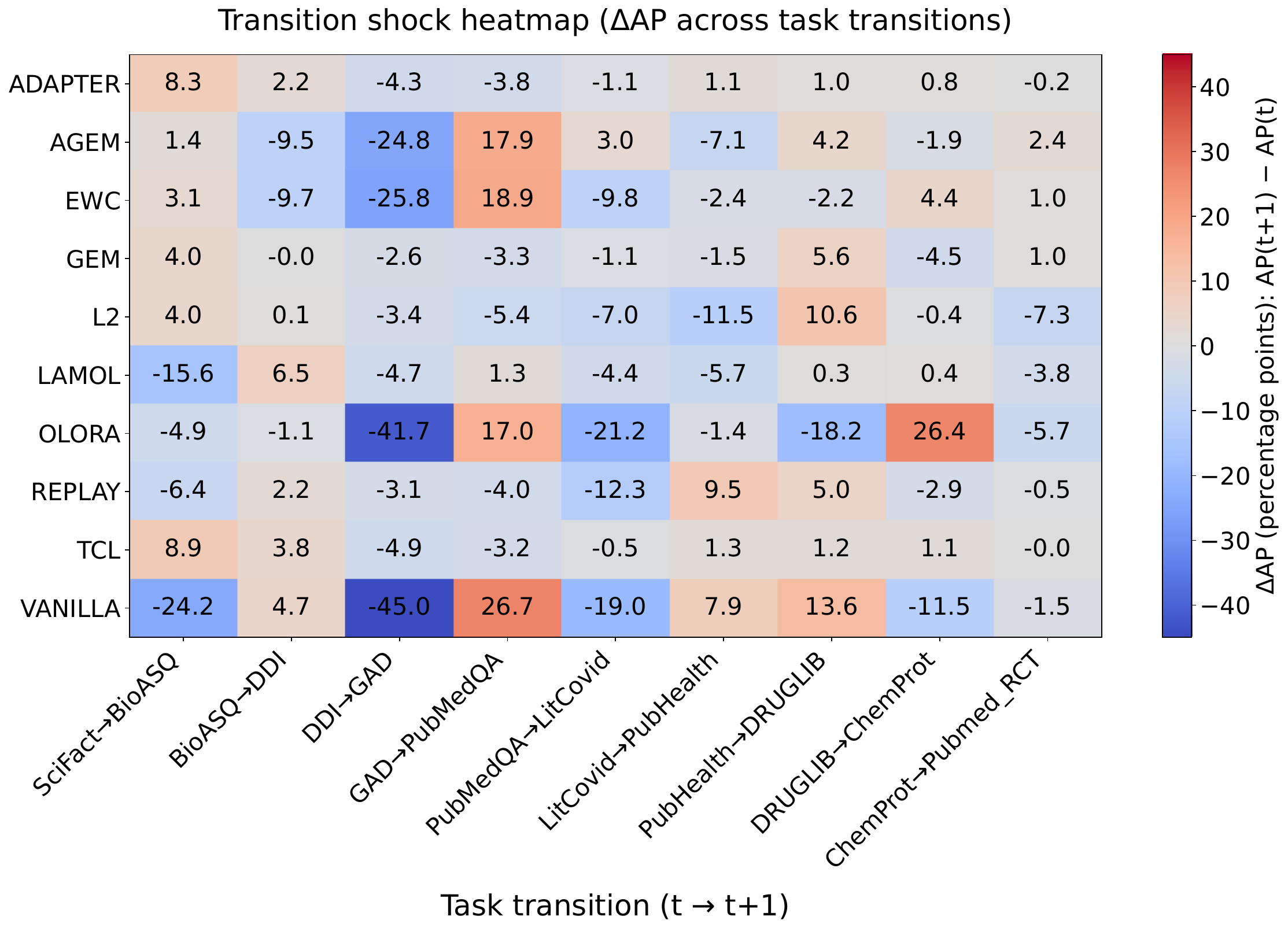}
    \caption*{Order~4}
  \end{subfigure}

  \caption{\textbf{Transition shock heatmaps under four task orders.}
  $\Delta AP_t = AP_{t+1}-AP_t$ (percentage points); cooler indicates interference.}
  \label{fig:order_dynamics_b}
\end{subfigure}
\end{figure}

\clearpage
\subsubsection*{Task-family differences in forgetting}
\begin{figure}[!htbp]
    \centering
    \includegraphics[width=\linewidth]{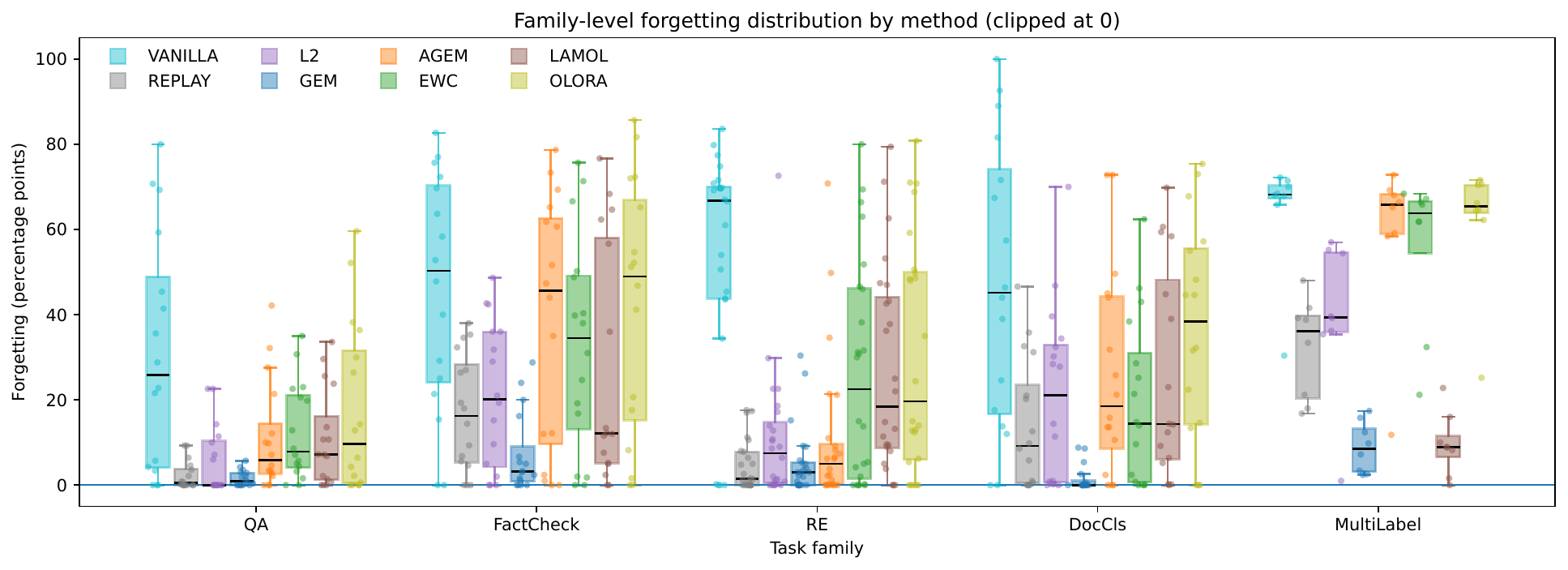}
    \caption{\textbf{Task-family forgetting distributions by method.}
    For each task, forgetting is measured as $\Delta = a_{t}^{\text{post}} - a_{t}^{\text{end}}$ (percentage points),
    i.e., the accuracy difference from immediately after learning the task to the end of the 10-task stream.
    Tasks are grouped into five families and distributions are aggregated across task orders.
    Boxes summarize the distribution over (order, task) instances and points show individual observations.
    We plot the clipped forgetting $\max(\Delta,0)$ to isolate performance loss without allowing gains (e.g., backward transfer) to offset forgetting.}
    \label{fig:family_forgetting_box_clip}
\end{figure}

To measure how much performance on each task degrades after subsequent updates, we compute per-task forgetting from the per-stage evaluations underlying Fig.~\ref{fig:order_dynamics_a}.
Specifically, forgetting on task $t$ is defined as $\Delta = a_{t}^{\text{post}} - a_{t}^{\text{end}}$ (percentage points).
Here, $a_{t}^{\text{post}}$ is the accuracy on task $t$ evaluated immediately after training on $t$ (i.e., after stage $t$).
$a_{t}^{\text{end}}$ is the accuracy on the same task evaluated after completing the full 10-task stream (i.e., after the final stage).
We then group the ten datasets into five task families using a fixed dataset-to-family mapping (Supplementary Note~\ref{supply:family_mapping})
and visualize the resulting forgetting distributions across task orders for a representative subset of methods (Fig.~\ref{fig:family_forgetting_box_clip}).
For clarity, we omit MULTI because it is trained jointly and does not yield a sequential trajectory required to define forgetting, and we omit ADAPTER and TCL because their task-specific parameter isolation can yield near-zero forgetting, which would compress the scale and obscure differences among non-isolation methods.

Across task orders, we observe clear family-level differences: MultiLabel tends to incur the largest forgetting, whereas \textsc{QA} and \textsc{RE} typically exhibit smaller losses, particularly for replay-based and other stability-oriented methods (e.g., REPLAY, GEM, L2).
Because $\Delta$ can be negative (e.g., backward transfer or evaluation variability), we use the clipped measure $\max(\Delta,0)$ as our primary forgetting metric,
and report the raw (unclipped) distributions in Supplementary Note~\ref{supply:task-family_raw}.

\subsection*{Scaling to LLMs}
\label{sec:scaling_llms}

We next examine whether our findings generalize and scale across backbone architectures, moving from an encoder--decoder model (\textsc{T5-base}) to decoder-only LLMs of different sizes.
We additionally evaluate two decoder-only LLMs, \textsc{Qwen-0.6B} and \textsc{Qwen-4B}.
Unless noted otherwise, scaling results use a single pre-specified task order (Order~1) and 5 epochs per task for tractability; the main benchmark uses 10 epochs per task (Table~\ref{tab:main_results}).
We use the same unified task formulation and evaluation protocol as in the main experiments to ensure comparability across backbones.

To contextualize scaling trends, Extended Data Table~\ref{edtab:scale_params} reports the final AP and the number of trainable parameters for each method under each backbone.
Fig.~\ref{fig:scale_cost_a} visualizes the corresponding method-wise AP after completing the 10-task stream.
Train counts parameters with \texttt{requires\_grad=True}.



\subsubsection*{Backbone scaling exhibits method-dependent and non-monotonic effects}
Extended Data Table~\ref{edtab:scale_params} and Fig.~\ref{fig:scale_cost_a} show that scaling effects are strongly method-dependent and can be non-monotonic.
Replacing T5 with a small decoder-only backbone (Qwen-0.6B) improves several baselines (e.g., VANILLA: 23.2$\rightarrow$41.43; EWC: 30.45$\rightarrow$44.98) but substantially degrades others (e.g., GEM: 68.83$\rightarrow$38.22; ADAPTER: 68.43$\rightarrow$59.67; TCL: 65.32$\rightarrow$57.26), indicating that architectural changes alone do not uniformly mitigate catastrophic forgetting.

Scaling further to Qwen-4B yields clearer improvements for many methods, particularly regularization-based approaches (EWC, L2) and replay-based training (REPLAY, LAMOL).
In contrast, the multi-task upper bound increases only modestly (MULTI: 75.08$\rightarrow$77.89), suggesting that additional capacity primarily reduces interference in the sequential setting rather than dramatically raising the joint-training ceiling.
Overall, backbone scaling reshapes the relative ranking of continual learning strategies rather than providing uniform gains across methods.


\subsubsection*{Parameter-efficient methods benefit unevenly from scaling}
Parameter-efficient approaches exhibit nuanced scaling behavior.
ADAPTER and TCL improve from Qwen-0.6B to Qwen-4B but do not recover the relative strength observed on T5, consistent with the fact that these methods freeze most backbone parameters and rely on limited task-specific capacity.
As the backbone scales, the fraction of trainable parameters becomes increasingly small, potentially turning the adapter module into a bottleneck that limits effective adaptation.
By contrast, the extremely compact OLORA shows only marginal improvement and remains substantially weaker than replay- or adapter-based methods, suggesting that aggressive parameter compression can undermine robustness to interference even on larger backbones.

\subsubsection*{Gradient-projection methods are highly architecture-sensitive}
Gradient-projection-based GEM exhibits pronounced sensitivity to backbone architecture.
While GEM is among the strongest methods on the encoder--decoder T5 backbone, its performance drops sharply on decoder-only models (38.22\% on Qwen-0.6B and 55.60\% on Qwen-4B), falling below replay- and regularization-based approaches on the same backbones.
A plausible explanation is that gradient-projection constraints can be more restrictive under decoder-only training dynamics, limiting effective updates compared to the encoder--decoder setting.
We leave a mechanistic analysis of this architecture sensitivity to future work.

\subsubsection*{Computational cost and efficiency trade-offs}
\begin{figure}[t]
    \centering

    \begin{subfigure}[t]{\linewidth}
        \centering
        \includegraphics[width=0.55\linewidth]{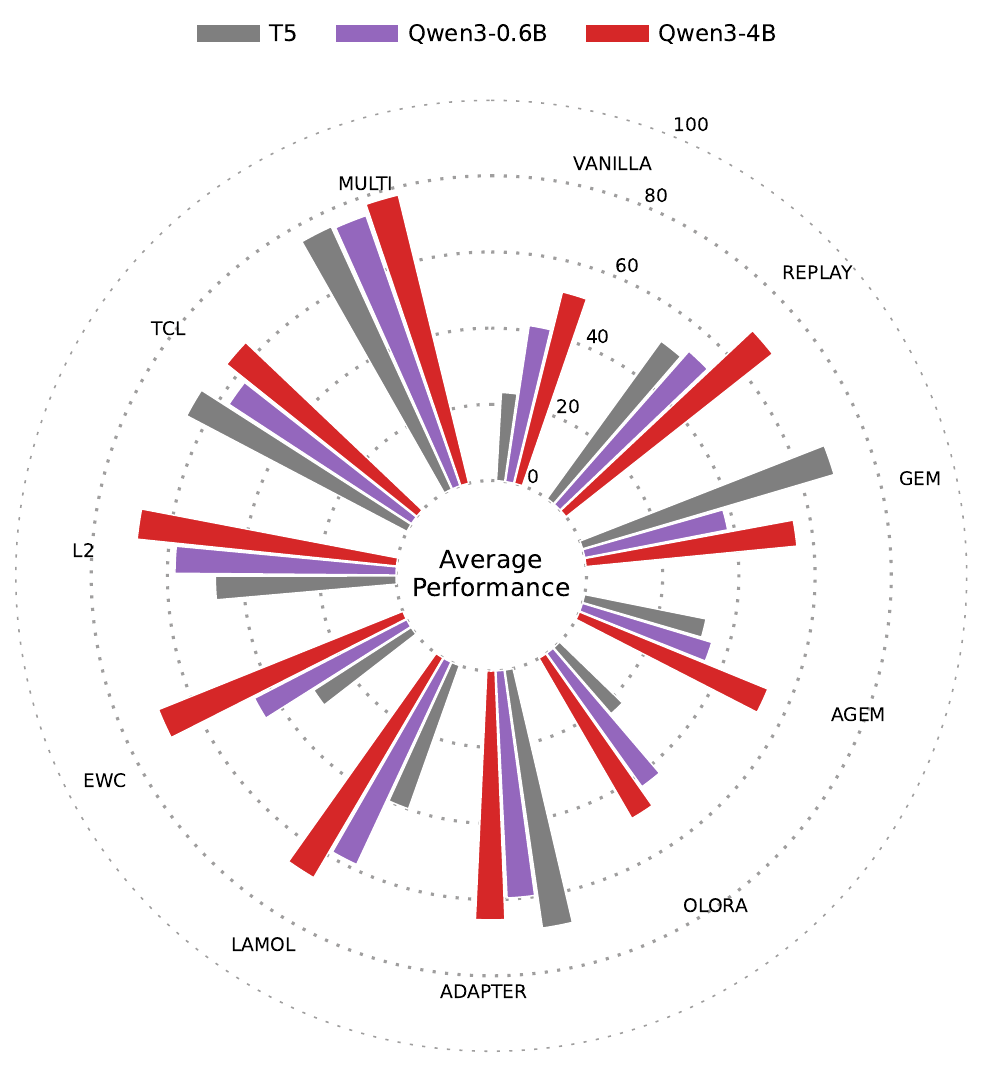}
        \caption{\textbf{Final AP across backbones.} }
        \label{fig:scale_cost_a}
    \end{subfigure}

    \vspace{0.5em}

    \begin{subfigure}[t]{0.48\linewidth}
        \centering
        \includegraphics[width=\linewidth]{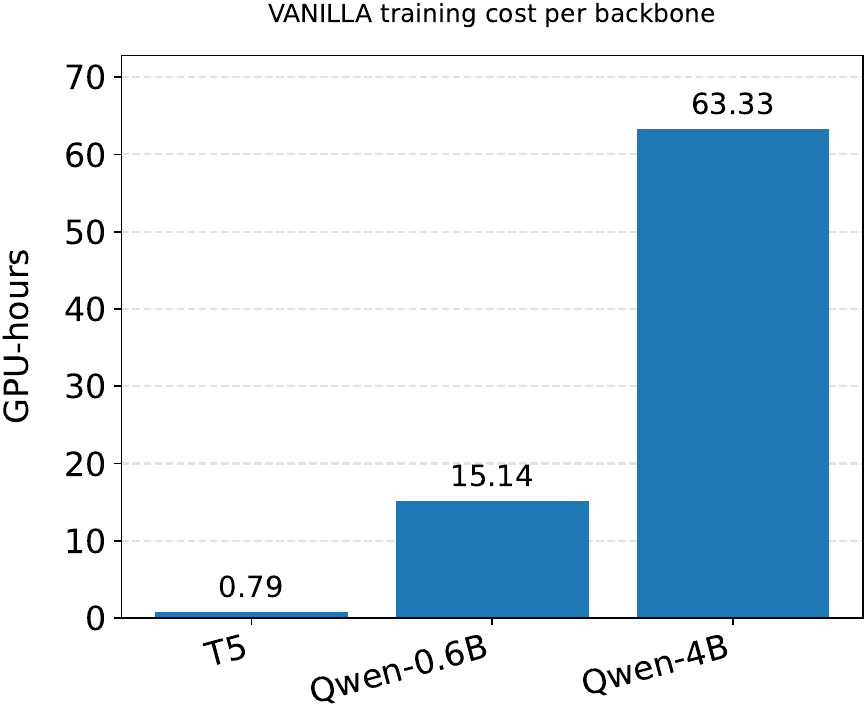}
        \caption{\textbf{VANILLA cost (GPU-hours).}}
        \label{fig:scale_cost_b}
    \end{subfigure}\hfill
    \begin{subfigure}[t]{0.48\linewidth}
        \centering
        \includegraphics[width=\linewidth]{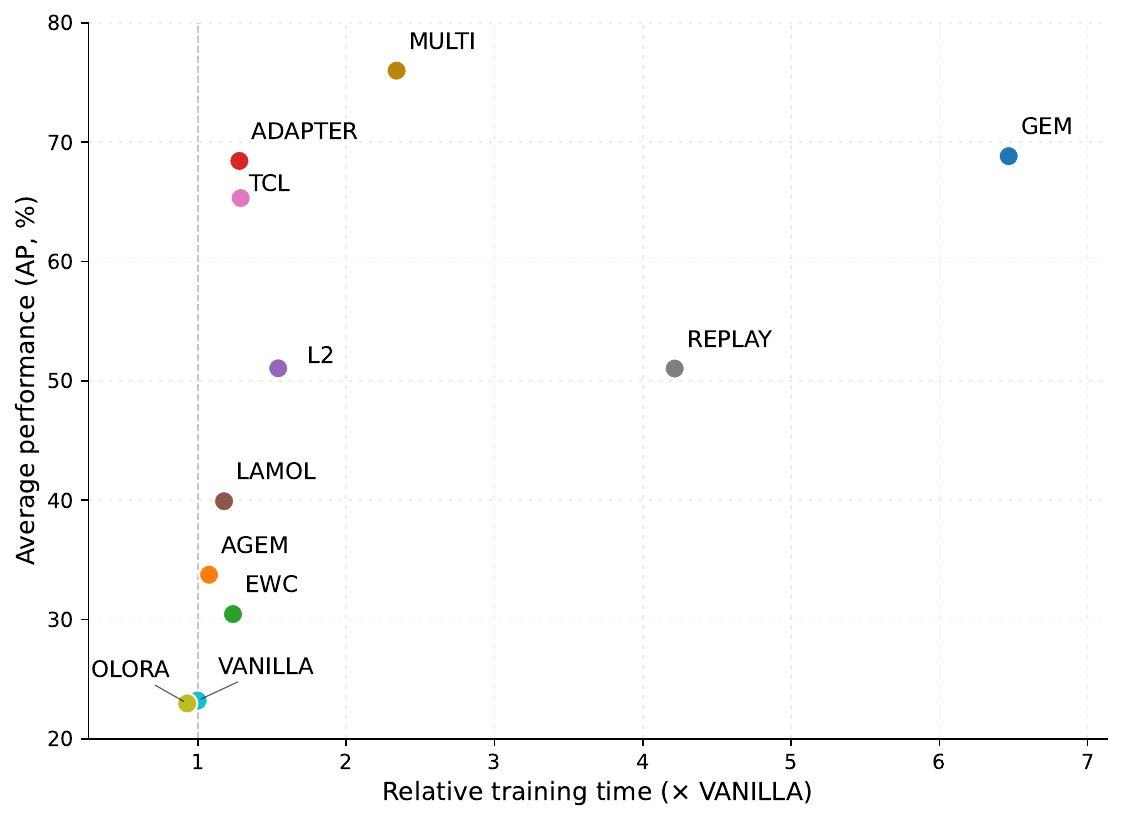}
        \caption{\textbf{\textsc{T5-BASE}: cost vs.\ AP} ($\times$ VANILLA time).}
        \label{fig:scale_cost_c}
    \end{subfigure}

    \vspace{0.5em}

    \begin{subfigure}[t]{0.48\linewidth}
        \centering
        \includegraphics[width=\linewidth]{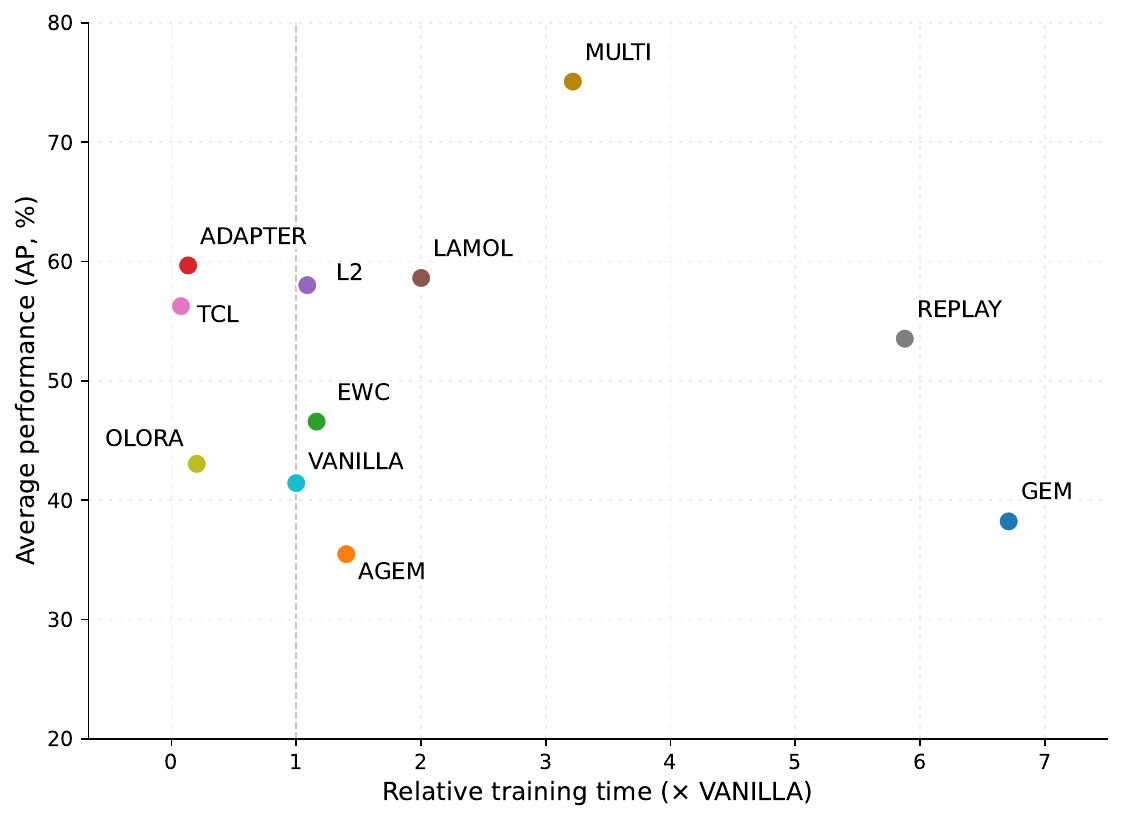}
        \caption{\textbf{\textsc{Qwen-0.6B}: cost vs.\ AP} ($\times$ VANILLA time).}
        \label{fig:scale_cost_d}
    \end{subfigure}\hfill
    \begin{subfigure}[t]{0.48\linewidth}
        \centering
        \includegraphics[width=\linewidth]{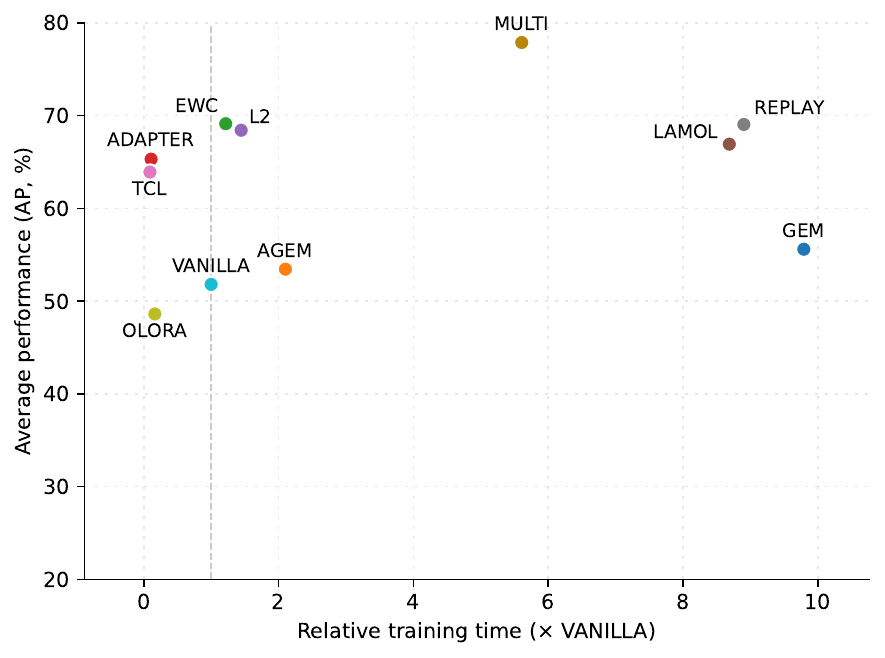}
        \caption{\textbf{\textsc{Qwen-4B}: cost vs.\ AP} ($\times$ VANILLA time).}
        \label{fig:scale_cost_e}
    \end{subfigure}
\caption{\textbf{Backbone scaling reshapes performance and efficiency trade-offs (Order~1).}
\textbf{(a)} Final AP after the 10-task stream on \textsc{T5-base}, Qwen-0.6B and Qwen-4B.
\textbf{(b)} Absolute training cost of VANILLA in GPU-hours for each backbone.
\textbf{(c--e)} AP versus relative training cost for \textsc{T5-base}, Qwen-0.6B and Qwen-4B, where cost is normalized by the corresponding VANILLA run on the same backbone.}
    \label{fig:scale_cost}
\end{figure}

We measure training cost in GPU-hours, computed as wall-clock time multiplied by the number of GPUs used.
Fig.~\ref{fig:scale_cost_b} reports the absolute training cost of VANILLA for each backbone, while
Figs.~\ref{fig:scale_cost_c}--\ref{fig:scale_cost_e} plot the average performance versus training-cost overhead normalized by the corresponding VANILLA run.
Full per-method training costs for each backbone are reported in Supplementary Note \ref{supply:training_time_backbone}.


MULTI incurs a moderate overhead relative to VANILLA, ranging from roughly $2\times$ on \textsc{T5-base} to $\sim\!6\times$ on \textsc{Qwen-4B}.
In contrast, replay-based methods are consistently more expensive due to interleaving memory samples, requiring roughly $4$--$9\times$ the GPU-hours of sequential fine-tuning across backbones.
On \textsc{T5-base}, GEM is also substantially costlier ($>6\times$), reflecting its additional per-step gradient constraints.

Taken together, these experiments reveal a nuanced scaling picture. 
(i) Moving from T5 to a small decoder-only LLM (\textsc{Qwen-0.6B}) does not automatically improve performance in biomedical NLP and can even degrade performance for some methods. 
(ii) Scaling further to \textsc{Qwen-4B} substantially mitigates catastrophic forgetting for many methods, particularly replay-based and regularization-based approaches. 
(iii) Algorithmic design remains crucial: replay-based methods and adapter-style parameter isolation provide consistently strong stability--efficiency trade-offs, whereas GEM shows pronounced backbone sensitivity in our experiments.


\clearpage
\clearpage

\section*{Discussion}
\label{sec:discussion}
In this study, we introduced MedCL-Bench, a continual learning benchmark that streams ten biomedical NLP datasets spanning five task families under a unified task formulation and evaluation protocol. 
Motivated by the need to update biomedical NLP models over time without retaining or retraining on all historical data, 
we use MedCL-Bench to characterize how CL strategies trade off adaptation to new tasks against retention of previously learned capabilities.
Across task-order permutations, VANILLA shows substantial regression on earlier tasks, whereas memory- and parameter-isolation strategies yield more stable behavior, with regularization-based baselines providing more limited protection.  More broadly, a central question is whether CL methods can reliably update biomedical models over evolving datasets while mitigating catastrophic forgetting, and which strategies offer the best trade-offs among end-of-stream performance, order robustness, parameter efficiency, and compute. MedCL-Bench is designed to address this question by evaluating methods across multiple task orders and complementary diagnostics that capture both end-of-stream outcomes and within-stream dynamics; we summarize the key findings below.

First, MedCL-Bench exposes large and systematic differences in stability across CL strategies.
Across eight randomized task orders, VANILLA exhibits severe catastrophic forgetting, whereas replay/constraint-based methods and parameter-isolation approaches substantially improve retention (Table~\ref{tab:main_results}). 
Forward transfer remains limited across methods on MedCL-Bench, underscoring that mitigating interference is the dominant challenge under long biomedical task streams.

Second, robustness to task-order permutations is a necessary reporting dimension rather than a secondary diagnostic.
Higher AP does not necessarily imply greater order robustness: some methods achieve competitive averages yet remain order-sensitive, whereas ADAPTER and TCL combine high performance with consistently low across-order variability (Fig.~\ref{fig:order_robustness}).
Several baselines show pronounced variability across orders, so conclusions from a single permutation can be unreliable; ADAPTER and TCL combine high average performance with the narrowest uncertainty intervals, indicating stronger order robustness.
Paired matched-order tests further show that key method differences persist across permutations under correction (Supplementary Note~\ref{supply:paired_tests}), supporting the use of task orders as matched blocks for statistical reliability.


Third, forgetting can be concentrated at specific task transitions. We quantify these drops with transition shock $\Delta AP_t$ (Fig.~\ref{fig:order_dynamics_b}), and larger or more frequent negative shocks are consistent with more negative BWT (Table~\ref{tab:main_results}). This transition-level view complements aggregate metrics by identifying when stability mechanisms succeed or fail during the stream.

Fourth, forgetting is heterogeneous across task families, highlighting clinically relevant failure modes that can be masked by overall averages.
When grouping datasets into five task families, we observe consistent family-level differences in forgetting, with some families exhibiting systematically larger loss across orders (Fig.~\ref{fig:family_forgetting_box_clip}). This heterogeneity suggests that benchmark reporting should include task-family diagnostics in addition to a single averaged score, especially in biomedical settings where failure on specific task types may be unacceptable even if the overall mean appears competitive.

Fifth, scaling to decoder-only LLMs reshapes the relative ranking of CL methods and reveals pronounced architecture dependence.
Moving from an encoder--decoder backbone (\textsc{T5-base}) to decoder-only backbones (\textsc{Qwen-0.6B/4B}) does not uniformly improve continual learning; instead, scaling effects are method-dependent and can be non-monotonic (Extended Data Table~\ref{edtab:scale_params}; Fig.~\ref{fig:scale_cost_a}). Regularization- and replay-based approaches benefit more consistently from increased decoder-only capacity, whereas parameter-efficient and gradient-projection methods exhibit stronger backbone dependence. 
In particular, GEM degrades on decoder-only models in our setting, indicating that the relative ranking of CL strategies can be backbone-dependent and motivating backbone-aware validation before deployment.

Sixth, stability gains come with a clear efficiency trade-off that should be reported alongside performance. Replay/constraint-based methods often deliver strong retention and lower transition shocks, but can incur substantial GPU-hour and memory overhead due to buffer maintenance and additional optimization constraints. In contrast, parameter-efficient approaches reduce the number of trainable parameters and can be more practical under limited compute, yet their benefits may saturate or become capacity-bottlenecked as backbones and task streams scale (Supplementary Note~\ref{supply:training_time_backbone}). These results highlight that method selection in biomedical settings should consider not only end-of-stream AP and robustness, but also the stability–compute–parameter trade-off under realistic update budgets.

Together, these results support two deployment-relevant recommendations. (i) When continual updates are expected, evaluate and select methods under multiple task orders and include uncertainty, since single-order reporting can be misleading. (ii) Choose methods based on both stability and compute: replay/constraint-based approaches offer strong retention but may incur substantial GPU-hour overhead, while parameter-efficient approaches reduce trainable parameters yet can become capacity-bottlenecked as backbones scale (Supplementary Note~\ref{supply:training_time_backbone}).

This study has several limitations. First, while the main benchmark evaluates multiple task orders, scaling experiments use a single representative order and fewer training epochs for tractability; extending scaling analyses to multiple permutations and longer horizons will strengthen generality. Second, our largest decoder-only model is limited to 4B parameters; evaluating larger biomedical LLMs may reveal additional scaling regimes. Third, we focus on accuracy under a unified task formulation; future work should incorporate additional deployment-facing criteria (e.g., calibration, robustness, or clinically weighted errors). Finally, exploring hybrid strategies that combine selective replay with parameter isolation, as well as principled memory sizing and sampling, may further improve the stability--compute trade-off.


In summary, MedCL-Bench fills a key gap in biomedical NLP by providing a unified continual learning benchmark that spans five task families, standardizes training and evaluation, and evaluates methods under multiple pre-specified task orders with complementary diagnostics. Using MedCL-Bench, we answer under-explored questions that motivate continual updating in practice: which CL strategies best prevent regressions under long biomedical task streams, how sensitive conclusions are to task-order permutations, and whether forgetting and interference differ systematically across task families. We further show that these conclusions depend on backbone architecture and are shaped by a clear stability--efficiency trade-off, motivating backbone-aware and budget-aware method selection. Overall, our findings indicate that CL outcomes are jointly governed by algorithm design, task-order permutations, task-family heterogeneity, and backbone architecture, providing practical guidance and more robust evaluation standards for updating biomedical language models under realistic constraints.



















\section*{Methods}\label{sec:methods}

\subsection*{Tasks and datasets}\label{sec:datasets}
MedCL-Bench benchmarks CL on ten publicly available biomedical NLP datasets spanning five task categories: biomedical question answering (BioASQ, PubMedQA), scientific fact checking (SciFact, PubHealth), relation extraction (GAD, ChemProt, DDI), document-level classification (PubMed-RCT, DRUGLIB), and multi-label topic classification of biomedical literature (LitCovid).
We adopt the released train/validation/test splits provided by the original datasets or the Zenodo benchmark release when applicable, and only construct missing splits when they are not provided.
For SciFact, the official test split is unlabeled; we therefore use the original validation split as the test and create a new validation split from the training data. 
For DRUGLIB, which provides train/test only, we create a validation split by sampling 10\% of the training set. 
For BioASQ (Task~7b), we retain only yes/no questions with gold answers in \{yes,no\}, use the first snippet as context, and create a deterministic train/validation split by hashing question IDs; the golden enriched set is used as test.
For PubMed-RCT, we apply light text cleaning and low-information filtering within each split, and (for controlled experiments) subsample up to 1,000 training sentences per section label.
For PubHealth, we convert the released parquet files into our unified JSONL format, keeping the claim text and the four-way labels (true/false/mixture/unproven) after filtering malformed entries.
For GAD, we keep the provided splits and rewrite each example into a unified binary MCQ format, where the input asks whether a gene--disease relation is present and provides two options (A: has\_relation, B: no\_relation); we map the original labels to A/B and keep the placeholder-marked text (e.g., @GENE, @DISEASE) as-is.
For the remaining datasets (PubMedQA, ChemProt, DDI, and LitCovid), we directly use the released train/validation/test splits from the Zenodo benchmark release (record 14025500) as-is.

\subsection*{Experimental setup}
\paragraph{Model backbones}
Unless stated otherwise, all main benchmark experiments use \textsc{T5-base} as the pretrained backbone (encoder--decoder).
For scaling experiments, we additionally evaluate two decoder-only backbones, \textsc{Qwen-0.6B} and \textsc{Qwen-4B}.

\paragraph{Continual learning protocol}
MedCL-Bench evaluates continual learning by exposing a pretrained backbone to a stream of $T{=}10$ tasks. 
Tasks are presented sequentially under eight pre-specified task-order permutations.
We fix the set of orders and use the same orders for all methods to enable matched comparisons across algorithms.
For the main benchmark results, we aggregate performance over the eight orders and report order sensitivity/uncertainty (e.g., s.d. and bootstrap CIs) to guard against order-specific conclusions.
For the LLM-scaling experiments, we use a single order (Order~1) for computational tractability, as detailed in the Results section (“Scaling to LLMs”).

\paragraph{Data curation and split caps}
To ensure a controlled and computationally consistent setting, we cap each dataset split on-the-fly to at most 1{,}000 training instances, 500 validation instances, and 500 test instances.
When a split exceeds the cap, we apply fixed-seed stratified subsampling within that split to preserve label proportions.
For datasets that do not provide all official splits (e.g., missing validation or a labeled test split), we construct the missing split(s) following the dataset-specific procedures described above.

\paragraph{Unified input/output format}
We cast all datasets into a unified discriminative classification format. 
Each instance is represented as a single text input (the \texttt{sentence} field), and the model predicts a label from a task-specific closed label set.
For QA tasks, the input provides a question and supporting context (and may optionally enumerate answer options), and the model predicts one answer label (BioASQ: \texttt{yes}/\texttt{no}; PubMedQA: \texttt{yes}/\texttt{no}/\texttt{maybe}). For relation and classification tasks (e.g., ChemProt, DDI, GAD, PubMed-RCT, DRUGLIB, SciFact, PubHealth), the model predicts one categorical label. For LitCovid, labels are multi-valued and represented as a semicolon-separated set; we evaluate predictions by exact set match (subset accuracy) after splitting labels by \texttt{;} in an order-insensitive manner.











\subsection*{Evaluation metrics}
MedCL-Bench spans heterogeneous biomedical task types whose original papers use different evaluation measures (e.g., accuracy, micro-/macro-F1). Directly aggregating such task-specific metrics would make CL summaries ill-defined and can introduce scale inconsistencies when averaging across tasks and task orders. We therefore cast all datasets into a unified classification setting and use accuracy as the primary metric, enabling consistent cross-task comparison and statistically coherent continual learning aggregates.
This design follows common practice in multi-task and continual learning benchmarks~\citep{madotto2021continual,zhang2022continual,zeng2026sparse} that enforce a shared output space (e.g., mapping heterogeneous tasks to a unified discriminative objective or to end-to-end generation) to support comparable aggregate measures.

All results are reported as percentages. For single-label tasks, an example is correct if the predicted label matches the gold label. For LitCovid (multi-label), we use subset accuracy (exact match): an example is counted as correct if and only if the predicted label set exactly matches the gold label set (order-insensitive, split by \texttt{;}).

Additionally, to obtain overall continual learning comparisons, we follow standard definitions~\citep{lopez2017gradient, madotto2021continual, zhang2022continual, zeng2026sparse} based on the task-wise accuracy matrix. Let $R_{t,i}$ denote the accuracy on task $i$ after training up to task $t$ (with $i \le t$) in the sequence, with $T$ total tasks.  In our benchmark, $T{=}10$.

\noindent(1) \textbf{Average Performance (AP)} is the average accuracy across all tasks after learning all the tasks: \[
\mathrm{AP}=\frac{1}{T}\sum_{i=1}^{T} R_{T,i}.
\]
(2) \textbf{Backward Transfer (BWT)}~\citep{lopez2017gradient, zhu2022continual} quantifies the impact of new learning on previous tasks:
\[
\mathrm{BWT}=\frac{1}{T-1}\sum_{i=1}^{T-1}\left(R_{T,i}-R_{i,i}\right).
\]
(3) \textbf{Forward Transfer (FWT)} measures the influence of previously learned tasks on future tasks before they are trained, relative to the initial pretrained (zero-shot) baseline:
\[
\mathrm{FWT}=\frac{1}{T-1}\sum_{i=2}^{T}\left(R_{i-1,i}-R_{0,i}\right),
\]
where $R_{0,i}$ denotes the performance on task $i$ of the initial pretrained model before learning any tasks (zero-shot baseline).

\subsection*{Baselines}\label{sec:baselines}
We compare representative baselines spanning sequential fine-tuning, multi-task learning, regularization, rehearsal/gradient-projection, and generative replay. All methods share the same backbone, preprocessing, task orders, and evaluation protocol.

\textit{Sequential fine-tuning (VANILLA)} trains on tasks one by one with standard fine-tuning and no explicit mechanism to mitigate forgetting.

\textit{Multi-task learning}: MULTI jointly trains on the union of all tasks and serves as a non-continual reference.

\textit{Regularization-based}: EWC~\citep{kirkpatrick2017overcoming} adds a Fisher-based penalty to prevent changing parameters important for previous tasks. L2 anchors parameters to the previous-task solution via an $\ell_2$ penalty.

\textit{Rehearsal/gradient projection}: REPLAY~\citep{robins1995catastrophic} maintains an episodic memory of past examples and mixes them with current-task data. GEM~\citep{lopez2017gradient} and AGEM~\citep{chaudhry2018efficient} use episodic memory to constrain updates and reduce interference with past tasks.

\textit{Generative replay}: LAMOL~\citep{sun2019lamol} performs replay by augmenting training with pseudo-samples generated from a model prior to learning the current task.

\textit{Parameter-efficient adaptation}: ADAPTER~\citep{madotto2021continual} inserts lightweight bottleneck adapters into the backbone and freezes all original backbone parameters, training only the adapter (and other explicitly enabled lightweight) parameters. 
TCL~\citep{zeng2025task} and OLORA~\citep{wang2023orthogonal} are additional parameter-efficient baselines that restrict training to a small set of task-conditioned lightweight parameters (e.g., task embeddings / low-rank updates) while keeping the backbone frozen, following their standard practice.  

\paragraph{Hyperparameter control}
Unless otherwise specified, we keep shared optimization hyperparameters (e.g., epochs, learning rate, batch sizes, early stopping) consistent across baselines for a given backbone. Method-specific hyperparameters (e.g., episodic memory size $M$ and LAMOL augmentation ratio $\rho$) are fixed across methods whenever they represent a shared resource budget, to ensure fair comparisons.

Hyperparameters: for rehearsal-based methods (REPLAY/GEM/AGEM), we use an episodic memory of size $M=5$ per task in our implementation. 
For LAMOL, we set \texttt{percentage\_LAM0L}{=}0.1, meaning that for each task we generate pseudo-samples amounting to 10\% of its training set using the model snapshot before learning that task.
For ADAPTER, we use a bottleneck size of 48 on T5 and update only adapter-related parameters.
For the T5-based ADAPTER, we use a bottleneck size of 48. 
For Qwen-based ADAPTER, we follow the same adapter-only training protocol and use a bottleneck size of 128/512 for 0.6B and 4B, respectively.
We use OLoRA with rank $r=8$ and $\alpha=16$.

\paragraph{Implementation details}
All methods are implemented in PyTorch with HuggingFace Transformers and DeepSpeed.
For each backbone, we use the same optimizer and training budget (epochs, learning rate, and batch sizes) across methods, and report results from the final checkpoint.
After finishing each task, we evaluate the model on all tasks seen so far to form the accuracy matrix $\{R_{t,i}\}$ used to compute AP/BWT/FWT.  The experiments were run on NVIDIA A100 40GB GPUs.

\paragraph{Use of large language models}
ChatGPT was used for language polishing; all scientific content and final text were authored and verified by the authors.

\section*{Data Availability}
All datasets used in this work are publicly available.
\begin{itemize}
    \item PubMedQA: \url{https://zenodo.org/records/14025500}
    \item BioASQ (Task 7b): the official training set (Training 7b) and test gold annotations (7b golden enriched) are available from the BioASQ Participants Area Datasets page (registration required for downloading the training set): \url{https://participants-area.bioasq.org/datasets/}
    \item PubHealth:\url{https://huggingface.co/datasets/ImperialCollegeLondon/health_fact}
    \item SciFact: \url{https://github.com/allenai/scifact} (also mirrored at \url{https://huggingface.co/datasets/allenai/scifact}).
    \item GAD:\url{https://huggingface.co/datasets/bigbio/gad}
    \item ChemProt: \url{https://zenodo.org/records/14025500}
    \item DDI: \url{https://zenodo.org/records/14025500}
    \item Pubmed\_RCT: \url{https://raw.githubusercontent.com/Franck-Dernoncourt/pubmed-rct/master/PubMed_20k_RCT}
    \item DRUGLIB: \url{https://archive.ics.uci.edu/dataset/461/drug\%2Breview\%2Bdataset\%2Bdruglib\%2Bcom}
    \item LitCovid: \url{https://zenodo.org/records/14025500}
\end{itemize}


\section*{Code Availability}
The code for MedCL-Bench, including dataset preprocessing scripts, continual learning configurations, and baseline implementations, has been deposited in Zenodo for peer review and will be made publicly available upon publication.

\bibliography{sn-bibliography}

\end{document}